\documentclass[runningheads]{llncs}
\usepackage{CJK}
\usepackage{tabularx}
\usepackage{multirow}
\usepackage[table,xcdraw]{xcolor}
\usepackage{colortbl} 

\usepackage[T1]{fontenc}
\usepackage{amsmath}

\usepackage{graphicx}
\usepackage{fancyhdr}
\usepackage{blindtext} % 仅用于生成示例文本

\newcommand{\shorttitle}{Joint Entity-Relation Extraction Model} % 文章的缩写标题
\newcommand{\authorname}{Feng et al}

\begin{document}

\title{The Joint Entity-Relation Extraction Model Based on Span and Interactive Fusion Representation for Chinese Medical Texts with Complex Semantics}
\author{ Danni Feng\inst{1} \and
Runzhi Li\inst{2} \and
Jing Wang\inst{1} \and
Siyu Yan\inst{3} \and
Lihong Ma\inst{3} \and
Yunli Xing\inst{4}}

\authorrunning{ Feng. Author et al.}
\institute{ School of Computer and Artificial Intelligence, Zhengzhou University\and
Cooperative Innovation Center of Internet Healthcare, Zhengzhou University \and
Fuwai Hospital, National Center for Cardiovascular Diseases, Chinese Academy of Medical Sciences and Peking Union Medical College\and
Department of Geriatrics, Beijing Friendship Hospital, Capital Medical University}

\maketitle

% 重新设置页眉样式，从第二页开始
\fancypagestyle{main}{
    \fancyhf{}
    \fancyhead[LE]{\thepage\ \quad \quad \authorname}
    \fancyhead[RO]{\shorttitle\ \quad \quad \thepage}
    \renewcommand{\headrulewidth}{0pt}
}
\pagestyle{main}
% typeset the header of the contribution
%
\begin{abstract}
Joint entity-relation extraction is a critical task in transforming unstructured or semi-structured text into triplets, facilitating the construction of large-scale knowledge graphs, and supporting various downstream applications. Despite its importance, research on Chinese text, particularly with complex semantics in specialized domains like medicine, remains limited. To address this gap, we introduce the CH-DDI, a Chinese drug-drug interactions dataset designed to capture the intricacies of medical text. Leveraging the strengths of attention mechanisms in capturing long-range dependencies, we propose the SEA module, which enhances the extraction of complex contextual semantic information, thereby improving entity recognition and relation extraction. Additionally, to address the inefficiencies of existing methods in facilitating information exchange between entity recognition and relation extraction, we present an interactive fusion representation module. This module employs Cross Attention for bidirectional information exchange between the tasks and further refines feature extraction through BiLSTM. Experimental results on both our CH-DDI dataset and public CoNLL04 dataset demonstrate that our model exhibits strong generalization capabilities. On the CH-DDI dataset, our model achieves an F1-score of 96.73\% for entity recognition and 78.43\% for relation extraction. On the CoNLL04 dataset, it attains an entity recognition precision of 89.54\% and a relation extraction accuracy of 71.64\%.

\end{abstract}
\section{Introduction}
Entity recognition and relation extraction are pivotal components in the establishment of large-scale knowledge graphs (KG). Lots of researches focus on them. One is two-stage pipeline method that trains two tasks independently\cite{mintz2009distant,zhong2020frustratingly,liu2023novel}. However, this approach encounters some significant challenges. Firstly, it tends to overlook the interplay between entity recognition and relation extraction. Secondly, it encounters error propagation. 

To solve the above problems, various joint entity relation extraction methods have been proposed. In which, traditional methods rely on labor-intensive and time-consuming feature engineering\cite{ren2017cotype,yu2010jointly}. Neural network-based joint entity relation extraction methods\cite{zheng-etal-2017-joint,ren-etal-2021-novel,shang2022onerel} demonstrate the capability to extract triplets in a single step. Nevertheless, lacking constraints on entity types, makes it difficult to guide relation prediction. Conversely, methods that leverage shared parameters offer flexibility, typically prioritizing entity recognition before relation extraction. However, they may overlook semantic information, which is particularly important for joint extraction of entities and relations, especially for Chinese texts. For instance, Zeng et al.\cite{zeng2014relation} employ Convolutional Neural Networks (CNN) to extract contextual information. Miwa et al.\cite{miwa2016end} and Zheng et al.\cite{zheng2017joint1} utilize Long Short-Term Memory (LSTM) networks to extract semantic information. However, These approaches struggle to effectively capture intricate details within sentences and grapple with capturing long-term dependencies. Subsequently, Eberts et al.\cite{Eberts2019SpanbasedJE} leverage the CLS token as a semantic indicator for entity recognition purposes. However, the CLS representation derived from BERT\cite{devlin-etal-2019-bert} may be susceptible to information loss or inadequate comprehension of the broader context. JI B et al.\cite{ji2020span} improved this work, achieving sentence-level semantic acquisition and span-level semantic acquisition through four different attention variants, but this also requires storing a large number of attention parameters.

Our research focuses on extracting structured knowledge from Chinese medical texts, which often contain complex semantics. To facilitate this, we constructed the CH-DDI, a drug-drug interactions dataset. When performing entity recognition, we integrate the advantages of existing models, maximize potential entity spans through span partitioning, and introduce attention-based SEA modules to effectively extract long-range contextual information. In addition, during the creation process of CH-DDI, we identified an important issue of relation overlap, where one drug may have the same or different interactions with multiple other drugs. Therefore, accurate semantic understanding during relation extraction is also crucial. To address this issue, we introduce local contextual semantics between span pairs through max-pooling during relation extraction.

Moreover, facilitating information exchange between the two tasks is deemed advantageous for enhancing model performance. Specifically, entity recognition by having awareness of the potential relations within a sentence can lead to more precise recognition of entities, and vice versa. For instance, given two potential location entities, the semantic relation in the sentence may be "Located\_in" rather than "Promotion". Xiong et al.\cite{xiong2022multi} demonstrated the usefulness of information exchange between tasks by implementing the gating mechanism for task interaction. Based on the idea that implementing information exchange between tasks can improve model performance, we propose an interactive fusion representation module to achieve bidirectional information exchange between entity recognition and relation extraction.

In this work, we propose a joint entity-relation extraction model based on span and interactive fusion representation for Chinese medical texts with complex semantics. In which, the interactive fusion representation module facilitates information exchange between entity recognition and relation extraction by Cross Attention and BiLSTM. And we use the span feature extraction module to obtain segmented spans and various features for spans. It should be noted that we treat the contextual semantics used in entity recognition as sentence-level semantics and use SEA for processing while treating the local context between span pairs used in relation extraction as spans. Similar to the extraction of internal features within spans, we use max-pooling to preserve the important features of the local context. The final entity recognition and relation extraction are the implementation of traditional classification.

Overall, our contributions are as follows:

1. We propose a joint entity-relation extraction model based on span and interactive fusion representation for Chinese medical texts with complex semantics, which captures long sentence dependencies through attention mechanism and extracts features again using BiLSTM, ultimately obtaining contextual semantics that consider both sentence importance and sentence sequence order.

2. We propose an interactive fusion representation module based on Cross Attention. The encoder in the model includes feedforward neural networks, which can obtain specific representations for entity recognition and relation extraction. The interactive fusion representation module can perform dual Cross Attention on these two representations and further use BiGRU for feature fusion.

3. We establish a Chinese drug-drug interactions dataset CH-DDI and conduct experiments on both CH-DDI and the public dataset CoNLL04. Compared with the baseline models, the best results are achieved on both Chinese and English datasets.

\section{Related Work} 
Entity recognition and relation extraction stand as pivotal sub-tasks within the realm of natural language processing(NLP). The initial two-stage model was gradually abandoned due to the difficulty in achieving interaction between two sub-tasks and the problem of error propagation. Currently, the joint extraction model for unified modeling the two tasks is used, and the extracted structured triplets are convenient for constructing large-scale knowledge graphs(KG) or completing other downstream tasks.    

\textbf{Parameter sharing-based methods} The shared-parameter joint extraction framework has been shown to enhance the performance of relation extraction through the synergistic optimization of entity recognition processes. Wei et al.\cite{wei-etal-2020-novel} utilizes the binary taggers to predict all potential head entities, and subsequently utilizes these predictions to infer their corresponding tail entities and relations. Fu et al.\cite{fu2019graphrel} innovatively constructs entities and relations as graphs, where entities are nodes and relations are edges. The method Bekoulis et al.\cite{bekoulis2018joint} proposed is similar to the work Huang et al.\cite{huang2019bert} proposed, they both use BIO annotation combined with CRF for entity recognition and allow for the prediction of multiple potential relations between a word and other words during relation extraction. The only difference lies in the selection of the basic encoder. However, the multi-head selection method lacks the use of semantic information for judgment. STER\cite{zhao2022exploring} is a model that uses knowledge distill, it has two teachers that study the information for entity recognition and relation extraction respectively. Our work, similar to SpERT\cite{Eberts2019SpanbasedJE} and SPAN\cite{ji2020span}, utilizes spans for overlapping entity recognition. However, neither of these methods can effectively achieve task interaction, and SpERT lacks effective acquisition of complex semantics. On the other hand, SPAN achieves sentence-level semantic supplementation through multiple attention mechanisms, which often require more storage of attention parameters.

\textbf{Joint decoding algorithm-based methods} The methods based on the decoding algorithm often blur the distinction between entity recognition and relation extraction, enabling the direct extraction of triplets in a single step. Common approaches that use joint decoders include tag-based and table-based. Zheng et al.\cite{zheng-etal-2017-joint} proposed a method that involves assigning a tag to each input token to denote the head or tail entities along with the relations, and subsequently inferring the triplet through tag prediction and parsing. However, the method based on tables often generates one table for each relation. Feiliang Ren et al.\cite{ren-etal-2021-novel} introduces a global feature-oriented relational triple extraction model, Wang et al.\cite{wang-etal-2020-tplinker} proposed a method that formulates joint extraction as a token pair linking problem and introduces a novel handshaking tagging scheme that aligns the boundary tokens of entity pairs under each relation type. Shang Y M et al.\cite{shang2022onerel} introduce a novel scoring based classifier to parallel tag and a Rel-Spec Horns Tagging strategy to ensure efficient decoding. Nevertheless, these methods often rely on predefined encoding and decoding strategies, and these methods often lack entity type constraints.

\textbf{Generative models-based methods} With the widespread application of generative models such as Transformer\cite{vaswani2017attention} in multiple fields, some researchers have also attempted to use generative models for entity relation extraction. Dianbo Sui et al.\cite{sui2023joint} used generative models to generate sets, where the elements in the set are the desired triplets. However, the output produced by the generative model is more stochastic, and this model often relies on richer training data. LinkNER\cite{zhang2024linkner} is the first work exploring how to fine-tune models synergistically with LLMs for NER tasks. Asada M et al.\cite{asada2024enhancing} Proposed a novel relation extraction method enhanced by large language models (LLMs).

\section{Method}
\subsection{Overview}
In this work, we propose a novel span-based joint entity-relation extraction model named ISER. It is illustrated in Figure 1. It primarily comprises five components that consist of the encoder module, the interactive fusion representation module, the span-based feature extraction module, the entity recognition module, and the relation extraction module. The encoder is a pre-trained BERT  followed by two feedforward layers.  The interactive fusion representation module carries out Cross Attention first and then executes BiLSTM to achieve representation fusion of entity and relation. The span-based feature extraction module implements span partition strategy, in which span width is embedded and span context is concated. The entity recognize module and relation extract module are traditional classification modules.

\begin{figure}
\includegraphics[width=\textwidth]{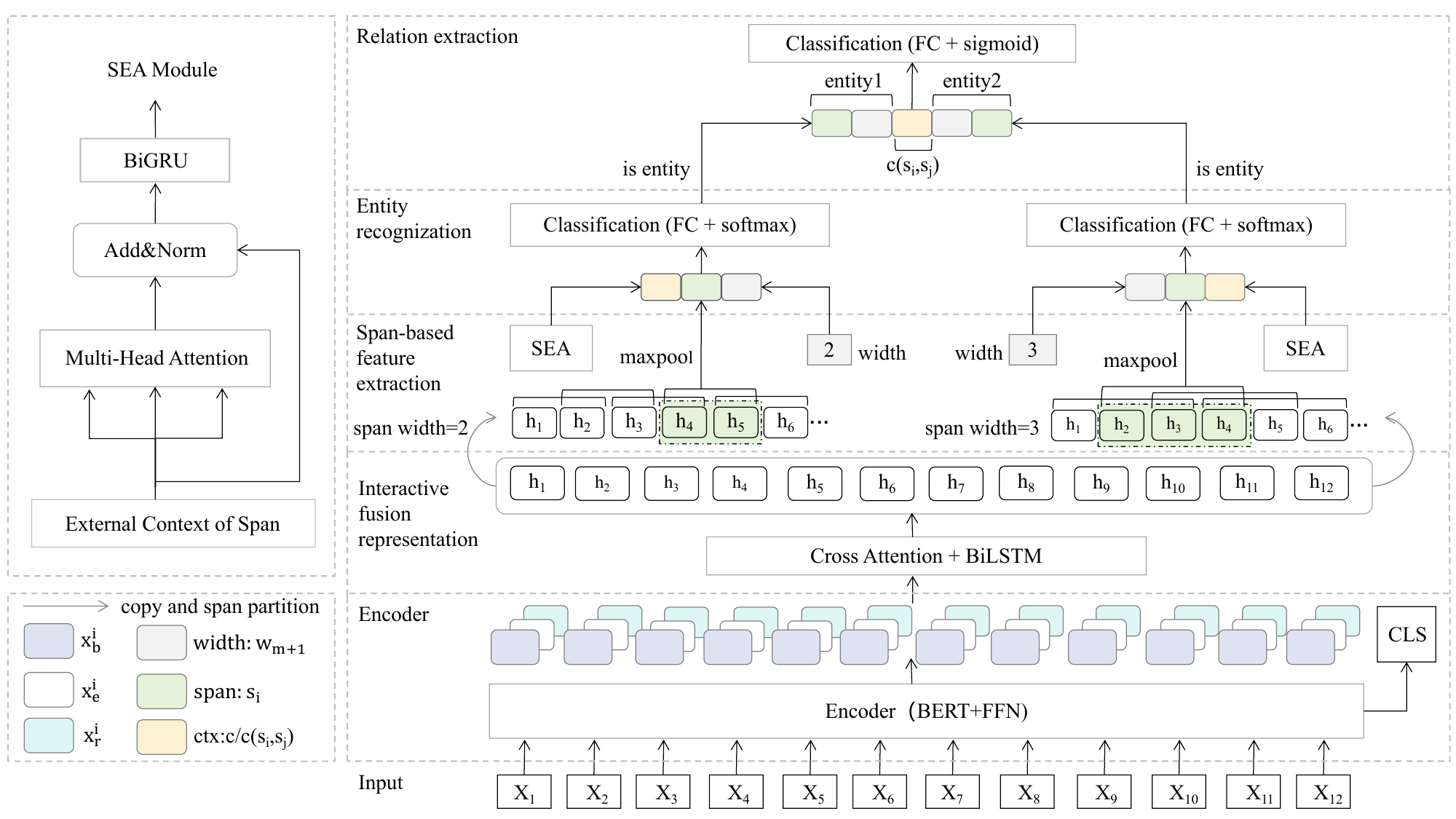}
\caption{The overall architecture of our model. The encoder can convert input sequences into three embedding representations: the embeddings obtained by BERT, representations for entity recognition and representations for relation extraction; the interactive fusion representation module can realize the information exchange between entity recognition and relation extraction; the span-based feature extraction module obtains three types features of spans: the internal features of spans, the width embedding and the external context; the entity recognition and relation extraction modules composed of traditional classifiers.} \label{fig1}
\end{figure}

\subsection{Encoder}
The encoder module is composed of a pre-trained BERT model and two feedforward neural networks (FFNs). The input sentence X is tokenized, obtaining a sequence of n tokens and passed through the encoder, obtaining an embedding sequence ($x_{b}^{1}$,$x_{b}^{2}$,$x_{b}^{3}$\ldots, CLS) of length n+1, along with specific embedding representations $X_{e}$($x_{e}^{1}$,$x_{e}^{2}$,\ldots $x_{e}^{n}$) and $X_{r}$($x_{r}^{1}$,$x_{r}^{2}$,\ldots $x_{r}^{n}$) of length n, which are derived through the FFN layers(Eq. 1,2) and respectively for entity recognition and relation extraction. In order to facilitate the completion of subsequent tasks, we map all features generated by the encoder to the same dimension as the BERT output and denote it as d.
\begin{equation}
 \mathrm{FFN\big(x_{b}^{i}\big)=\big(ReLU\big(x_{b}^{i} W_{e}^{1}+b_{e}^{1}\big)\big)W_{e}^{2}+b_{e}^{2}}
\end{equation}
\begin{equation}
     \mathrm{FFN\big(x_{b}^{i}\big)=\big(ReLU\big( x_{b}^{i}W_{r}^{1}+b_{r}^{1}\big)\big)W_{r}^{2}+b_{r}^{2}}
\end{equation}
Both W and b are trainable weights and biases.
\subsection{Interactive fusion representation}
The employment of the attention mechanism facilitates the selective focus on critical information within input sequences, and Cross Attention allows for the exchange of information across disparate data, thereby enhancing the model's capacity to comprehend semantic context. As outlined in the introduction, we are supposed to consider the impact of entity recognition on relation extraction and vice versa. Consequently, the integration of Cross Attention is proposed to promote mutual interaction between entity recognition and relation extraction. After the interactive information is concatenated, it is further fused through BiLSTM. The architecture of the interactive fusion representation is depicted in Fig 2.
\begin{figure}
\includegraphics[width=\textwidth,height=7cm]{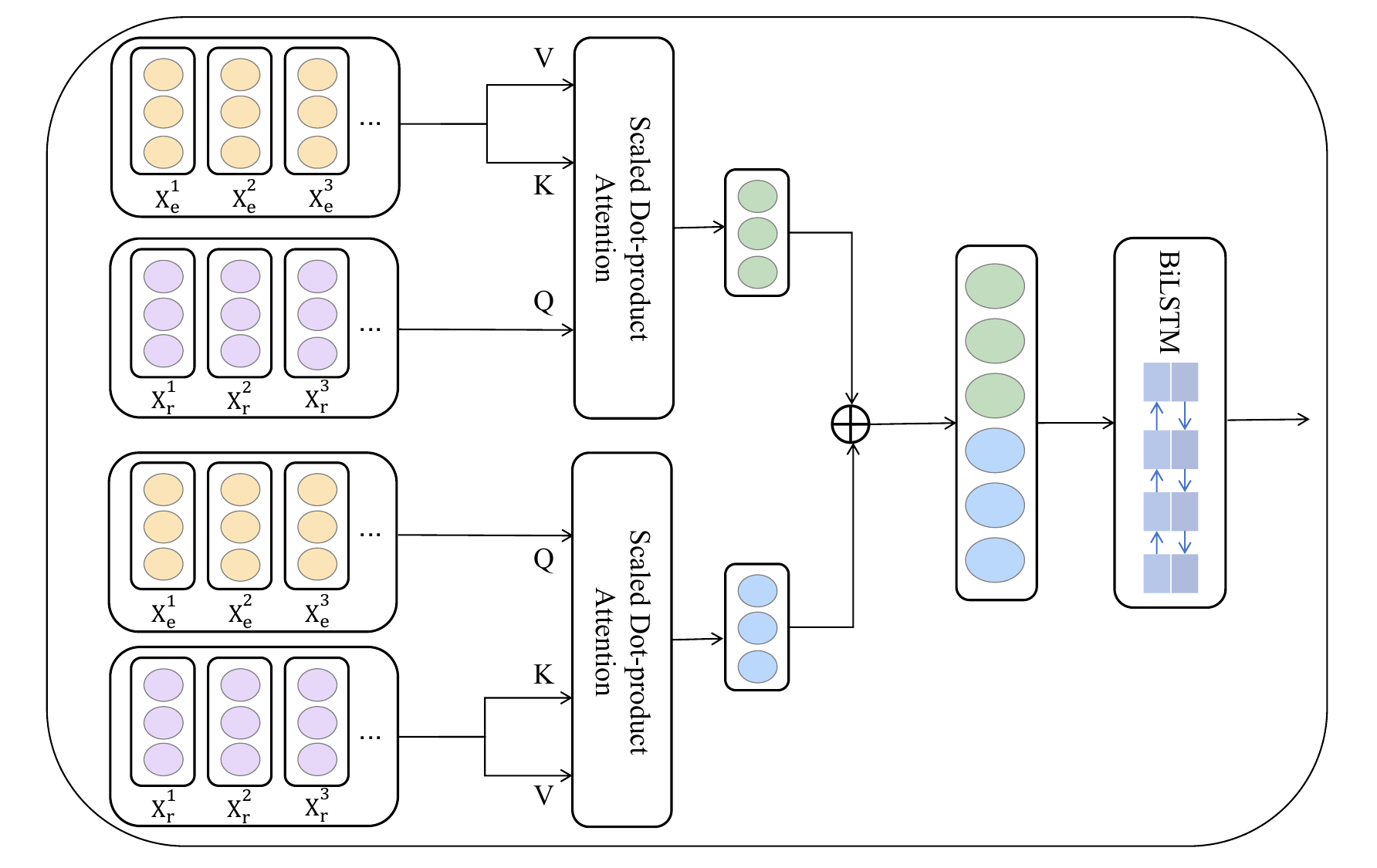}
\caption{The overall architecture of the interactive fusion representation.} \label{fig2}
\end{figure}

In this work, we use the scaled dot-product attention mechanism to implement Cross Attention. Specifically, using one of the representations obtained by the encoder for entity recognition($X_{e}$) and relation extraction($X_{r}$) as Query and the other as Key and Value. Through the computation of the scaled dot-product operation (Eq. 3,4), revised representations for both entity recognition and relation extraction are derived. Subsequently, these revised representations $\widetilde{X}_{e}$ and $\widetilde{X}_{r}$ undergo concatenation via a Bidirectional Long Short-Term Memory (BiLSTM) architecture (Eq. 5), resulting in a novel embedding sequence H = ($h_{1}$,$h_{2}$,...$h_{n}$). This newly formed representation, reflecting the latest state post bilateral information exchange, is harnessed for the execution of subsequent tasks.
\begin{equation}
    \widetilde{X}_{e}=\mathrm{softmax}\left(\frac{\mathrm{X_{r}X^{T}_{e}}}{\sqrt{\mathrm{d}}}\right)\mathrm{X_{e}}
\end{equation}
\begin{equation}
    \widetilde{X}_{r} =\mathrm{softmax}\left(\frac{\mathrm{X_{e}X^{T}_{r}}}{\sqrt{\mathrm{d}}}\right)\mathrm{X_{r}}
\end{equation}
\begin{equation}
    H=BiLSTM(\widetilde{X}_{e}\oplus\widetilde{X}_{r})
\end{equation}
$H$ is the final result obtained by the interactive fusion representation module, and each token $h_{i}$ in it is a d-dimensional vector.
\subsection{Span-based feature extraction}
In our span-based feature extraction module, a slide window strategy is employed to facilitate effective span partition from input sequences. Consider an input sequence of length n, where the sliding window is configured with a size k. This approach systematically traverses the sequence, extracting all candidate spans of size k, illustrated in Fig 3(a). We constrain the window size because named entities tend not to span excessively long sequences, thereby enhancing computational efficiency. For each candidate span generated, a dual mask technique is applied to encode both the span and its contextual information. The encoding of span is that positions occupied by span tokens are assigned 1 while assigning 0 to all non-span positions; Inversely, the context mask assigns span token positions with 0 and all other context positions with 1, as depicted in Fig 3(b).
\begin{figure}
\includegraphics[width=\textwidth]{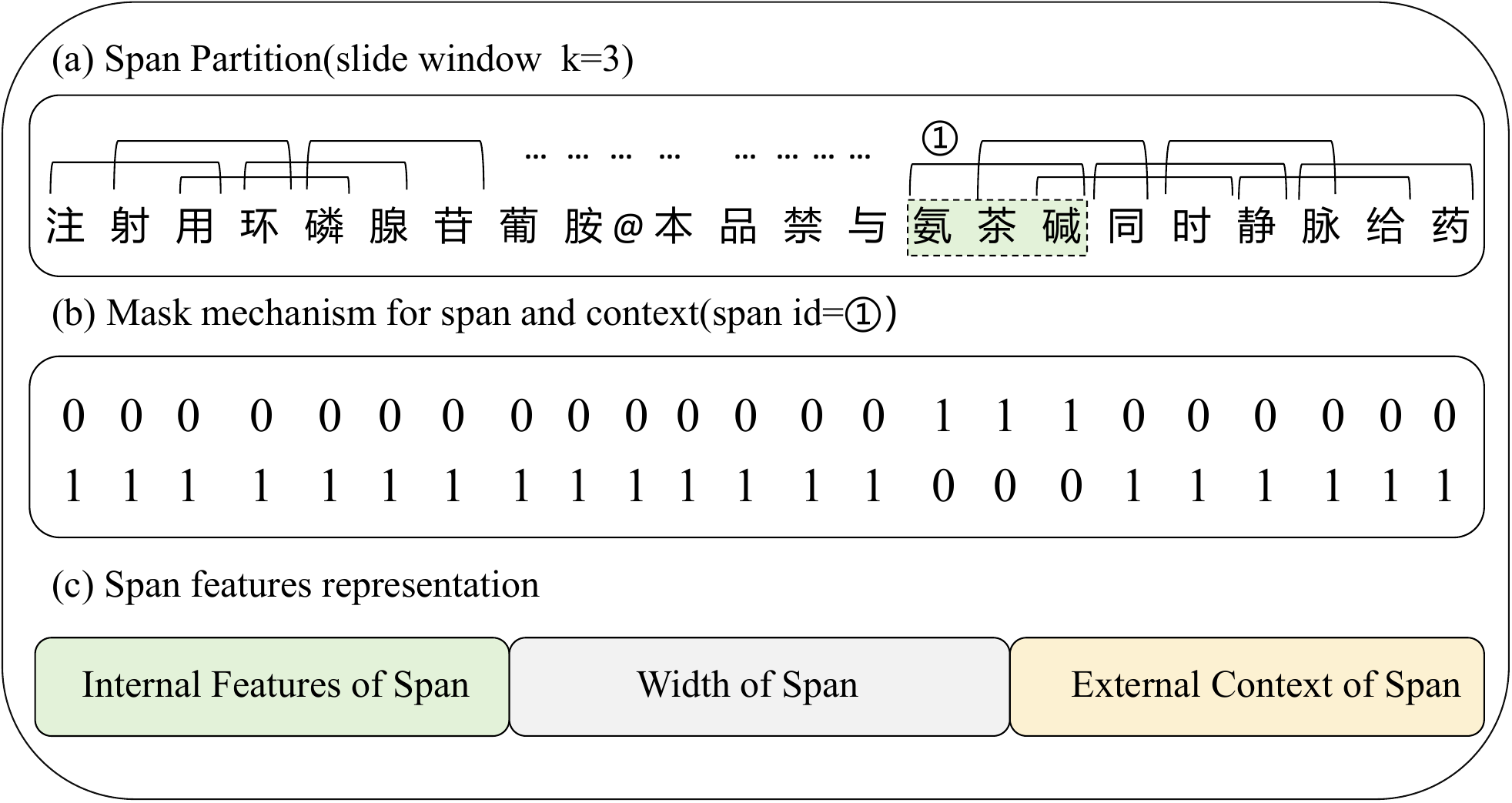}
\caption{The overall process of feature extraction based on span.} \label{fig3}
\end{figure}

After partitioning the span, we extract three types of span features separately. For the existing sequence H = ($h_{1}$,$h_{2}$,...$h_{n}$) obtained through interactive fusion representation module and the span s =($h_{i}$,$h_{i+1}$,$h_{i+2}$) with width of 3 to be predicted, we use max-pooling to extract internal features across spans(Eq. 6); We then use word embedding to map the discrete span width to a fixed dimensional continuous vector, and this feature is denoted as $w_{3}$ for span s and is used to clarify the boundary information of the span. Finally, We use the proposed SEA module for context feature extraction, which consists of Multi-Head Attention and BiGRU, as shown in Fig 4. The input of the SEA module is the sequence with the span has been masked $H_{mask}$=($h_{1}$,...$h_{i-1}$,[mask],[mask],[mask],$h_{i+3}$,..$h_{n}$). Then the Multi-head Self-attention is used to capture contextual semantic information. To ensure the integrity of the original information, residual structures are introduced. Meanwhile, due to the temporal insensitivity of the attention mechanism, a Bidirectional Gated Recurrent Unit (BiGRU) network is integrated into the SEA architecture. The final output of the SEA module is denoted as c. For inputs with dimension d, the calculation formulas of the SEA module are as follows (Eq. 7,8,9).
\begin{equation}
    \mathrm{s~=~maxpool}(h_{\mathrm{i}},h_{\mathrm{i+1}},h_{\mathrm{i+2}})
\end{equation}
\begin{equation}
    K/Q/V=H_{mask}W_{K/Q/V}
\end{equation}
\begin{equation}
\text{Attention(K,Q,V)}=\text{softmax}\left(\frac{\mathrm{QK^T}}{\sqrt{\mathrm{d}}}\right)\text{V}
\end{equation}
\begin{equation}
    \mathrm{c=BiGRU(Attention(K,Q,V))}
\end{equation}
$W_{K},W_{Q},W_{V}$ are trainable weights.

In Fig 3(c), the three types of span features are concatenated and then used for entity recognition.

\begin{figure}[htbp!]
\includegraphics[scale=0.5]{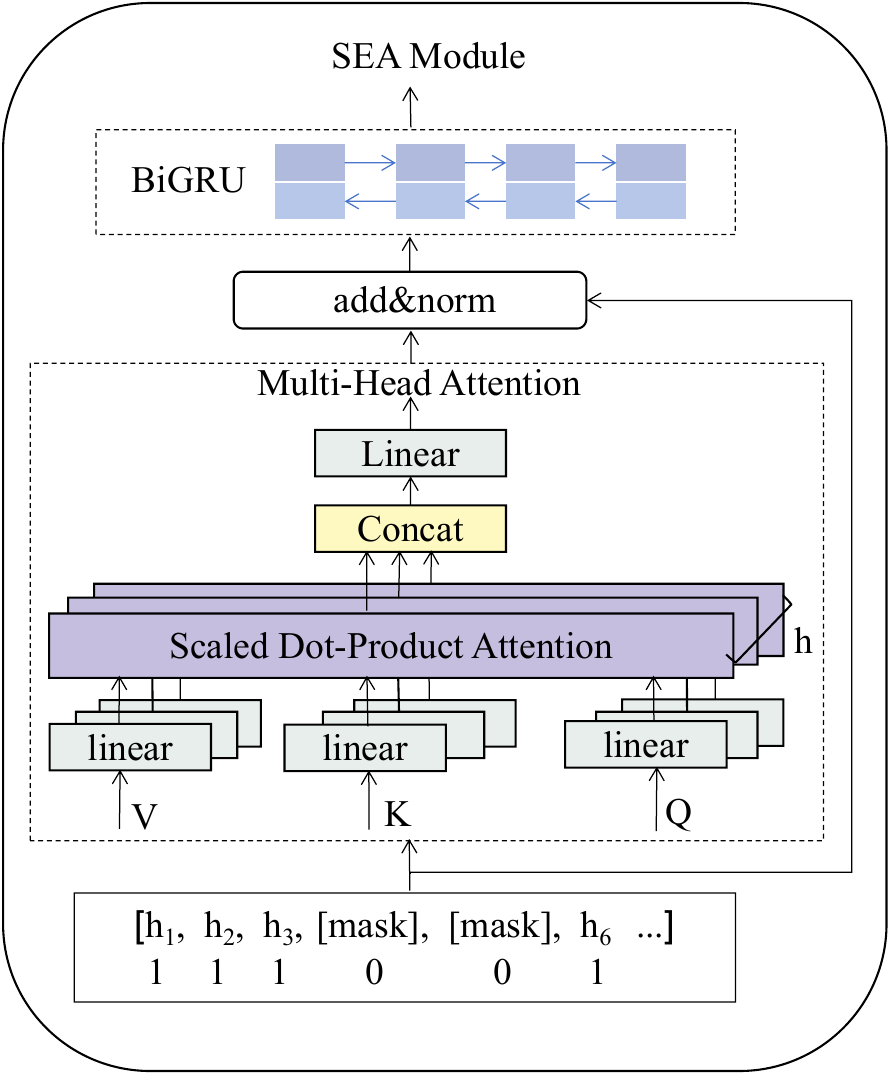}
\centering
\caption{The architecture of the SEA module.} \label{fig3}
\end{figure}
\subsection{Entity recognition}
Our entity recognition classifier is a multi-class classifier and it uses any candidate spans as input. For example, let span =($h_{i}$,$h_{i+1}$,...$h_{i+m}$) represent a span whose width is m+1. We let $\mathbf{\varepsilon}$ represent the set of entity types included in the dataset, the classifier can map the span to an entity class. During the entity recognition phase, the representations we use are the concatenated representations obtained from the span-based feature extraction module $z_{e}$. 

\begin{equation}
    \mathrm{z_{e}=s\oplus w_{m+1}\oplus c}
\end{equation}
\begin{equation}
    \widetilde{y}_e=softmax\big(W_e^{'}\cdot z_e+b_e^{'}\big)
\end{equation}
\subsection{Relation extraction}
The relation extraction classifier uses candidate span pairs and the local context between them as input to predict whether there is a relation. For the given span pair (s$_1$, s$_2$) whose widths are p,q respectively, the span pair (s$_1$, s$_2$) are represented by their internal features and width features, and we can easily obtain contextual information between span pairs because we save mask records of spans when partitioning them, and we defined the context between the span pairs as c(s$_1$, s$_2$). As relations are often asymmetric, we will predict whether there is a relation both in (s$_1$, s$_2$) and (s$_2$, s$_1$). It should be noted that we also consider the local context between span pairs as "spans", and like obtaining internal features of spans, we use max-pooling to obtain contextual information.

\begin{equation}
\mathrm{z_r^1=(s_1\oplus w_p)\oplus c(s_1,s_2)\oplus(s_2\oplus w_q)}
\end{equation}
\begin{equation}
   \mathrm{z_r^2=(s_2\oplus w_q)\oplus c(s_2,s_1)\oplus(s_1\oplus w_p)}
\end{equation}
\begin{equation}
    \widetilde{y}_r^{1/2}=\sigma\left(W_r^{'}\cdot z_r^{1/2}+b_r^{'}\right)
\end{equation}
\subsection{Training}
We use text annotated with entities and relations for training. By training, the fine-tuning BERT, a width embedding matrix for spans, and classifiers for entity recognition and relation extraction can be obtained. The loss of our model we defined is as follows (Eq. 15).
\begin{align}
    \mathrm{L=L_{e}+L_{r}}
\end{align}

The L$_{e}$ is the entity recognition loss and the L$_{r}$ is the relation extraction loss. In the context of entity recognition, the model's output necessitates processing through a softmax activation function to yield the entity type with the maximum probability. The cross-entropy loss function (Eq. 16) is employed to compute the discrepancy between the predicted and actual categorical distributions for entity labels. Conversely, for relation extraction, the task reduces to determining whether a specific relation exists or not, which can be framed as a binary classification problem. Thus, the sigmoid function is utilized to map the model's output to a probabilistic representation confined between 0 and 1, and subsequently, the binary cross-entropy loss function is applied to measure the loss incurred during this binary classification process (Eq. 17). ($y_{e}$ and $y_{r}$ are the Ground Truth)
\begin{align}
    &\mathrm{L_e}=CrossEntropyLoss(\widetilde{y}_{e},y_{e})\\
    &\mathrm{L}_{\mathrm{r}}= BCEWithLogits(\widetilde{y}_{r}, y_{r})
\end{align}
\section{Experiments}
\subsection{Datasets}
\subsubsection{CoNLL04:} The CoNLL04 dataset\cite{carreras-marquez-2004-introduction} is a widely-used public benchmark for joint entity and relation extraction, it comprises four primary entity types (Location, Organization, Person, and Other) and five distinct relation types (Work\_for, Kill, Live\_in, Organized\_in, and Located\_in). In this paper, We conducted experiments using the dataset provided by SpERT, which includes 922 training datasets and 231 testing datasets.
\vspace{-6mm}
\begin{CJK}{UTF8}{gbsn}
\subsubsection{CH-DDI:} This is a self-built Chinese drug-drug interactions dataset, which uses drug information from the website Yimaitong. Specifically, the steps for constructing the dataset are as follows: (1) Crawl drug-drug interactions in drug instrcutions related to five chronic diseases (coronary heart disease, COPD (Chronic Obstructive Pulmonary Disease), diabetes, hypertension, and osteoporosis), remove useless information, and obtain the original data after preliminary processing. Taking the treatment of hypertension with bumetanib tablets as an example, the raw data obtained is ``<布美他尼片><药物相互作用><1.与多巴胺合用，利尿作用加强。2.肾上腺糖、盐皮质激素，促肾上腺皮质激素及雌激素能降低本药的利尿作用。3.与拟交感神经药物及抗惊厥药物合用，利尿作用减弱>''. (2) Perform sentence segmentation to obtain sentences containing complete drug interactions. After this processing, the above example will obtain ``布美他尼片@与多巴胺合用，利尿作用加强||
布美他尼片@肾上腺糖、盐皮质激素，促肾上腺皮质激素及雌激素能降低本药的利尿作用||
布美他尼片@与拟交感神经药物及抗惊厥药物合用，利尿作用减弱'' (3) Refer to the annotation of CoNLL04 to achieve annotation of our dataset. Taking the first sentence after sentence segmentation as an example, the annotated data obtained is ``\{"text": "布美他尼片@与多巴胺合用，利尿作用加强", "spo\_list": [{"predicate": "协同", "object\_type": {"@value": "药物"}, "subject\_type": "药物", "object": {"@value": "布美他尼片"}, "subject": "多巴胺"}]\}''. In summary, under the guidance of clinical doctors, drug interactions are defined as seven types (antagonism, inhibition, synergy, taboo, irrelevance, promotion, addition), and we only marked drug entities with significant interactions. The processing flow of the entire dataset follows the above process and has been checked by three people.
\end{CJK}
The statistical results of the two datasets introduced above are shown in Table 1, And the statistical of the relations in the CH-DDI are presented in Table 2.
\begin{table}[htbp]
\centering
\caption{The statistic of CoNLL04 and CH-DDI}
\begin{tabularx}{0.8\textwidth}{p{0.1\textwidth}>{\centering\arraybackslash}X>{\centering\arraybackslash}X>{\centering\arraybackslash}X>{\centering\arraybackslash}X>{\centering\arraybackslash}X>{\centering\arraybackslash}X} 
\hline
      & \multicolumn{3}{c}{CoNLL04} & \multicolumn{3}{c}{CH-DDI} \\ \hline
      & sentence & entity & relation & sentence & entity & relation \\ \hline
train & 922      & 3377   & 1283     & 585      & 1830   & 1276     \\
test  & 231      & 893    & 343      & 147      & 527    & 386      \\ \hline
\end{tabularx}
\end{table}

\begin{table}[]
\centering
\caption{The Statistic of Relations in CH-DDI}
\begin{tabularx}{0.7\textwidth}{p{0.2\textwidth}>{\centering\arraybackslash}X>{\centering\arraybackslash}X>{\centering\arraybackslash}X>{\centering\arraybackslash}X>{\centering\arraybackslash}X>{\centering\arraybackslash}X} 
\hline
Relation types & Training dataset & Testing dataset \\ \cline{1-3} 
Antagonism  & 252              & 76              \\
Inihibition & 120              & 22              \\
Synergy     & 319              & 118             \\
Taboo       & 279              & 91              \\
Irrelevance & 187              & 31              \\
Promotion   & 95               & 42              \\
Addition    & 24               & 6               \\ \hline
\end{tabularx}
\end{table}
\subsection{Evaluation Metrics and experiment settings}\label{SCM}
\textbf{Evaluation Metrics:} This article assesses the performance of the model in entity recognition and relation extraction. For entity recognition, a prediction is considered correct when a span is identified as an entity and its boundaries and types exactly match the ground truth. As for relation extraction, it is deemed accurate if both spans are correctly recognized as entities and the relation between these spans is accurately predicted. In particular, we compute the precision (Eq. 18), recall(Eq. 19), and F1-score(Eq. 20) on both macro and micro indicators for both entity recognition and relation extraction. 

\begin{align}
    &\text{Precision}=\frac{\mathrm{TP}}{\mathrm{TP+FP}}\\
    &\mathrm{Recall}=\frac{\mathrm{TP}}{\mathrm{TP}+\mathrm{FN}}\\
    &\mathrm{F1}=\frac{2*\mathrm{Precision}*\mathrm{Recall}}{\mathrm{Precision}+\mathrm{Recall}}
\end{align}

\subsubsection{Experiment Settings:} The model architecture incorporates BERT$base$ as the underlying encoder for the CoNLL04 dataset, while Bio-BERT, specifically pre-trained on medical text, is utilized for the CH-DDI dataset. Both BERT models undergo fine-tuning during the training process to adapt to the respective tasks. Training involves the implementation of the Adam optimizer with a learning rate decay schedule, commencing with a peak learning rate of 2e-5. To optimize computational efficiency and convergence properties, the batch size for training iterations is empirically determined to be 2 following extensive experimental validations. Subsequently, the batch size is adjusted to 1 during the testing phase. Through meticulous experimentation, a relation threshold of 0.4 is established for the relation extraction. Based on Figure 5, we set the dropout for the CoNLL04 to 0.1 and for the CH-DDI, it is 0.75. The epoch for training we set is 70. The same method as Eberts was used to generate negative samples, 100 negative entity and relation samples were generated during the training for each sentence in CoNLL04 and 50 negative entity and relation samples for each sentence in CH-DDI. 
\begin{figure}
\includegraphics[width=\textwidth]{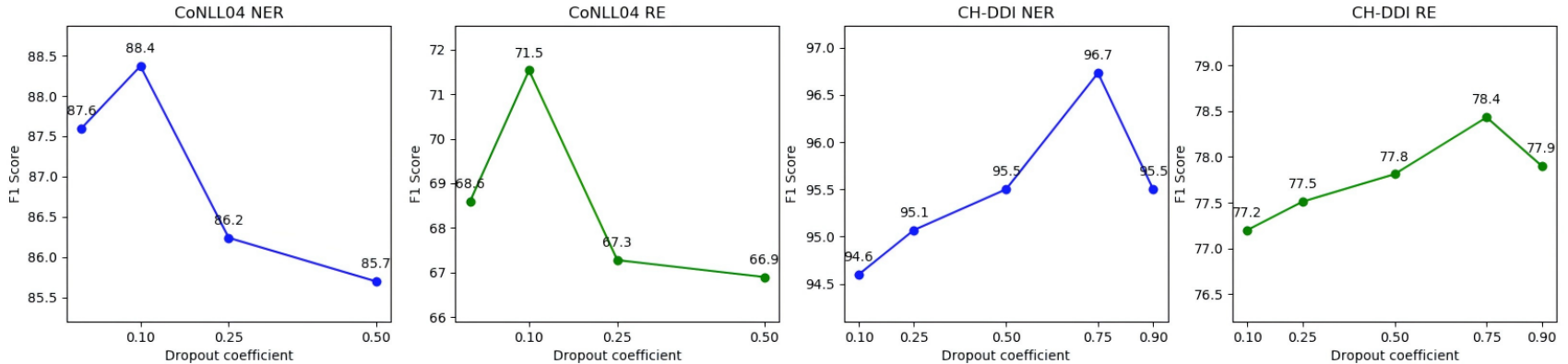}
\centering
\caption{Effect of dropout coefficient on CoNLL04 and CH-DDI.} \label{fig3}
\end{figure}
\subsection{Comparison with baseline models}

Table 3 and Table 4 shows the comparison results between our model and other models on two datasets. To ensure the reliability of our comparison, we reproduced all the benchmark models and conducted experiments on the two datasets introduced above to obtain and compare the best results.

\textbf{GraphRel\cite{fu2019graphrel}:} It is an end-to-end relation extraction model that uses graph convolutional networks (GCNs) to jointly learn named entities and relations.

\textbf{CasRel\cite{wei-etal-2020-novel}:} This model utilizes binary taggers to extract entities and relations. It needs to pre-define the relation patterns (subject\_type, relation, object\_type).

\textbf{SPN4RE\cite{sui2023joint}:} SPN4RE is a prediction model designed for triple sets, which does not require consideration of the sequential output of traditional seq2seq models.

\textbf{Multi-head\cite{bekoulis2018joint}:} It uses the CRF layer for entity recognition and then uses a Multi-head selection mechanism in the relation extraction.

\textbf{SpERT\cite{Eberts2019SpanbasedJE}:} This model is a span-based joint extraction method, which uses an entity classifier and a relation classifier to obtain entities and relations.

\textbf{OneRel\cite{shang2022onerel}:} It is based on decoding algorithms and obtains relation triplets in one step. 

\textbf{STER\cite{zhao2022exploring}:} It is a joint extraction model with knowledge distill framework and it has two teacher models for entity recognition and relation extraction respectively.

\textbf{W2NER\cite{lu2022unified}:} This model is based on tables, which generates a table for each input sequence and decodes the table to obtain the predicted entity and entity type.

\textbf{ISER$_{GRU}$:} It refers to remove the Semantic Enhancement Attention(SEA) module and only uses GRU to extract context information.

\textbf{ISER$_{ChineseBERT}$:} It refers to uses Chinese\_BERT for CH-DDI rather than the Bio-BERT which has trained on large biomedical materials.
\begin{table}[]
\centering
\caption{Compares our model with other models on the CH-DDI}
\begin{tabularx}{\textwidth}{p{0.2\textwidth}>{\centering\arraybackslash}X>{\centering\arraybackslash}X>{\centering\arraybackslash}X>{\centering\arraybackslash}X>{\centering\arraybackslash}X>{\centering\arraybackslash}X} 
\hline
\multirow{2}{*}{Methods} & \multicolumn{3}{c}{NER}                          & \multicolumn{3}{c}{RE}                           \\ \cline{2-7} 
                         & P              & R              & F              & P              & R              & F              \\ \hline
GraphRel$_{1p}$\cite{fu2019graphrel}                 & 51.98          & 68.85          & 59.23          & 31.76          & 49.48          & 38.69          \\
GraphRel$_{2p}$\cite{fu2019graphrel}                 & 53.19          & 63.35          & 57.83          & 36.42          & 47.64          & 41.31          \\
SPN4RE\cite{sui2023joint}                   & 85.54          & 81.61          & 83.53          & \textbf{84.04} & 80.17          & \textbf{82.06} \\
Multi-head\cite{bekoulis2018joint}               & 89.03          & 91.67          & 90.33          & 69.21          & 70.83          & 70.01          \\
SpERT\cite{Eberts2019SpanbasedJE}                    & 92.63          & 97.72          & 95.11          & 75.38          & 76.96          & 76.17          \\
OneRel\cite{shang2022onerel}                   & -              & -              & -              & \textbf{78.89} & 78.27          & \textbf{78.58} \\
STER\cite{zhao2022exploring}                     & 94.09          & 96.64          & 95.35          & 74.27          & 80.10          & 77.08          \\
W2NER\cite{lu2022unified}                    & 95.05          & 96.04          & 95.54          & -              & -              & -              \\
ISER$_{GRU}$                      & 94.11          & 97.53          & 95.79          & 74.87          & 78.01          & 76.41          \\
ISER$_{ChineseBERT}$                      & 94.01          & 98.29          & 96.10          & 75.25          & 79.58          & 77.35          \\
ISER(micro)              & \textbf{95.22} & \textbf{98.17} & \textbf{96.73} & \textbf{76.11} & \textbf{80.89} & \textbf{78.43} \\
ISER(macro)              & \textbf{98.40} & \textbf{99.43} & \textbf{98.90} & \textbf{77.24} & \textbf{83.09} & \textbf{79.30} \\ \hline
\end{tabularx}
\end{table}
\begin{table}[]
\centering
\caption{Compares our model with other models on the CoNLL04}
\begin{tabularx}{\textwidth}{p{0.2\textwidth}>{\centering\arraybackslash}X>{\centering\arraybackslash}X>{\centering\arraybackslash}X>{\centering\arraybackslash}X>{\centering\arraybackslash}X>{\centering\arraybackslash}X} 
\hline
                          & \multicolumn{3}{c}{NER}                                                                    & \multicolumn{3}{c}{RE}                                                                     \\ \cline{2-7} 
\multirow{-2}{*}{Methods} & P                            & R                            & F                            & P                            & R                            & F                            \\ \hline
GraphRel$_{1p}$\cite{fu2019graphrel}                  & 41.28                        & 48.98                        & 44.80                        & 36.90                        & 47.23                        & 41.43                        \\
GraphRel$_{2p}$\cite{fu2019graphrel}                  & 44.39                        & 48.40                        & 46.30                        & 40.78                        & 45.77                        & 43.13                        \\
CasRel\cite{wei-etal-2020-novel}                    & -                            & -                            & -                            & 58.95                        & 43.83                        & 50.28                        \\
SPN4RE\cite{sui2023joint}                    & 63.67                            & 57.73                            & 60.55                            & 63.02                        & 57.14                        & 59.94                        \\
Multi-head\cite{bekoulis2018joint}                & 83.75                        & 84.06                        & 83.90                        & 63.75                        & 60.43                        & 62.04                        \\
SpERT\cite{Eberts2019SpanbasedJE}                     & 86.69                        & 86.79                        & 86.74                        & 72.31                        & 64.72                        & 68.31                        \\
OneRel\cite{shang2022onerel}                    & -                            & -                            & -                            & 68.62                        & 70.64                        & 69.62                        \\
STER\cite{zhao2022exploring}                      & 88.88                        & 86.79                        & 87.82                        & \textbf{75.00}               & 64.72                        & 69.48                        \\
W2NER\cite{lu2022unified}                     & 84.86                        & 85.18                        & 85.02                        & -                            & -                            & -                            \\
ISER$_{GRU}$                      & 86.99                        & 87.57                        & 87.28                        & 71.47                        & 66.47                        & 68.88                        \\
ISER(micro)               & \textbf{89.54}               & \textbf{87.23}               & \textbf{88.37}               & \textbf{71.64}               & \textbf{71.43}               & \textbf{71.53}               \\
IESR(macro)               & \textbf{86.66}               & \textbf{82.44}               & \textbf{84.32}               & \textbf{72.92}               & \textbf{73.42}               & \textbf{72.77}               \\ \hline
\end{tabularx}
\end{table}
% \usepackage{multirow}

\begin{CJK}{UTF8}{gbsn}
 Comparative analysis of evaluation metrics reveals the necessity of leveraging domain-specific pre-training models such as the use of Bio-BERT for the medical field, because the Bio-BERT can provide some potential knowledge. As shown in Section 4.1, the annotation data our self-built dataset is:"\{"text": "布美他尼片@与多巴胺合用，利尿作用加强", "spo\_list": [{"predicate": "协同", "object\_type": {"@value": "药物"}, "subject\_type": "药物", "object": {"@value": "布美他尼片"}, "subject": "多巴胺"}]\}. Before the "@" symble, it is an entity involved in drug interactions, which can be seen as entity prompting. And the dataset only contains one entity type。 Therefore, the performance of entity recognition on this dataset is excellent. Moreover, from the statistical situation of various relation types in Table 2, there is a data imbalance problem in the CH-DDI dataset, which may have a certain impact on relation extraction.
 \end{CJK}
% \usepackage{multirow}

 Our model demonstrates superior performance in joint extraction tasks. Experiments with the Graphrel\cite{fu2019graphrel} and Casrel\cite{wei-etal-2020-novel} models indicate their reliance on substantial training data, the small-sized dataset we use leading to subpar performance. SPN4RE\cite{sui2023joint} is a seq2seq model designed for end-to-end task, simultaneously generating labels for entities and their corresponding triples. As a result, the model maintains consistent performance across both entity recognition and relation extraction. Our approach resembles a pipeline model, it concentrates on the two tasks for entities and relations, including improving performance in both tasks to ensure broad applicability. The multi-head selection\cite{bekoulis2018joint} model excels in entity recognition but falls short in relation extraction due to limited semantic information. Shang et al.\cite{shang2022onerel} proposed the one-stage triplet extraction model OneRel. It can effectively address the problem of relation overlap and has shown good performance on both datasets we use. STER is a model that uses knowledge distill to joint extract entities and relations which has excellent performance. In contrast to the span-based SpERT\cite{Eberts2019SpanbasedJE} model, our model achieves superior performance in both entity recognition and relation extraction. On the CH-DDI dataset, the F1-score of our model for both tasks increases by 1.62\% and 2.26\%. On the CoNLL04 dataset, our model exhibits improvements of 1.63\% and 3.22\% in the F1-score for entity recognition and relation extraction, respectively. The W2NER\cite{lu2022unified} achieves comprehensive entity recognition across domains and languages, showcasing superior performance. However, from Table 3 and Table 4, it can be seen that our model performs better than W2NER on both the CH-DDI dataset and the CoNLL04 dataset, with F1-score improved by 1.19\% and 3.35\% respectively.
\vspace{-5MM}

\subsection{Ablation study}
We performed seven ablation experiments, the results are presented in Table 5 and Table 6. In these experiments, ``-interactive fusion'' refers to the removal of the interactive fusion representation module; ``SEA -$>$ CLS'' denotes the elimination of the proposed SEA module and the utilization of the CLS token generated by BERT; ``LC -$>$$LC_{att}$'' means we use self-attention to obtain the local context; ``LC -$>$$LC_{BERT}$'' refers to using the local context in the original BERT output; ``LC -$>$ CLS'' signifies the utilization of the CLS token output by BERT; ``LC -$>$$GC_{BERT}$'' refers to using the whole output generated by BERT and ``LC -$>$ GC'' denotes utilizing the whole output obtained through interactive fusion representation module.
\begin{table}[]
\centering
\caption{The results of the ablation study on CH-DDI}
\begin{tabularx}{\textwidth}{p{0.25\textwidth}>{\centering\arraybackslash}X>{\centering\arraybackslash}X>{\centering\arraybackslash}X>{\centering\arraybackslash}X>{\centering\arraybackslash}X>{\centering\arraybackslash}X} 
\hline
                                                                              & \multicolumn{3}{c}{NER}                          & \multicolumn{3}{c}{RE}                           \\ \cline{2-7} 
                                                                              & P              & R              & F              & P              & R              & F              \\ \hline
\begin{tabular}[c]{@{}l@{}}-interactive fusion \end{tabular} & 94.03          & 98.67          & 96.30          & 74.88          & 79.58          & 77.16          \\
SEA-\textgreater{}CLS                                                         & 94.11          & 96.96          & 95.51          & 74.62          & 77.75          & 76.15          \\
LC-\textgreater{}LC$_{att}$                                                           & 94.98          & 97.83          & 96.38          & 76.92          & 79.11          & 78.00          \\
LC-\textgreater{}LC$_{BERT}$                                                           & 94.81          & 97.15          & 95.97          & 76.84          & 78.13          & 77.48          \\
LC-\textgreater{}CLS                                                          & 94.83          & 97.53          & 96.16          & 75.14          & 78.80          & 76.93          \\
LC-\textgreater{}GC$_{BERT}$                                                           & 94.68          & 97.91          & 96.27          & 76.08          & 78.27          & 77.16          \\
LC-\textgreater{}GC                                                           & 95.18          & 97.34          & 96.25          & 77.00          & 78.01          & 77.50          \\
ISER                                                                          & \textbf{95.22} & \textbf{97.72} & \textbf{96.73} & \textbf{76.11} & \textbf{80.89} & \textbf{78.43} \\ \hline
\end{tabularx}
\end{table}
\vspace{-5mm}
 \begin{table}[]
\centering
\caption{The results of the ablation study on CoNLL04}
\begin{tabularx}{\textwidth}{p{0.25\textwidth}>{\centering\arraybackslash}X>{\centering\arraybackslash}X>{\centering\arraybackslash}X>{\centering\arraybackslash}X>{\centering\arraybackslash}X>{\centering\arraybackslash}X} 
\hline
                                                                              & \multicolumn{3}{c}{NER}                          & \multicolumn{3}{c}{RE}                           \\ \cline{2-7} 
                                                                              & P              & R              & F              & P              & R              & F              \\ \hline
\begin{tabular}[c]{@{}l@{}}-interactive fusion \end{tabular} & 87.40          & 87.01          & 87.21          & 70.59          & 66.47          & 68.47          \\
SEA-\textgreater{}CLS                                                         & 87.77          & 87.75          & 87.67          & 68.19          & 69.39          & 68.79          \\
LC-\textgreater{}LC$_{att}$                                                           & 88.55          & 88.02          & 88.28          & 70.78          & 69.93          & 70.35          \\
LC-\textgreater{}LC$_{BERT}$                                                           & 88.46          & 87.57          & 88.01          & 73.62          & 65.89          & 69.54          \\
LC-\textgreater{}CLS                                                          & 88.98          & 87.68          & 88.32          & 60.44          & 71.72          & 65.60          \\
LC-\textgreater{}GC$_{BERT}$                                                           & 87.90          & 86.23          & 87.05          & 61.52          & 66.18          & 63.76          \\
LC-\textgreater{}GC                                                           & 88.94          & 86.45          & 87.67          & 66.28          & 66.47          & 66.38          \\
ISER                                                                          & \textbf{89.54} & \textbf{87.32} & \textbf{88.37} & \textbf{71.64} & \textbf{71.43} & \textbf{71.53} \\ \hline
\end{tabularx}
\end{table}

Based on the experimental results presented in Table 5 and Table 6, it can be observed that our proposed interactive fusion representation module effectively enhances model performance, particularly in relation extraction, with an increase of 1.27\% and 3.06\% in F1-score on the CH-DDI and CoNLL04 datasets, respectively. After removing the Semantic Enhancement Attention (SEA) module we proposed and replacing it with the token CLS, the performance of entity recognition on both datasets decreased, which in turn affected subsequent relation extraction. 

In addition, we focused on discussing the semantic information used in the relation extraction, the Fig 6 illustrates the comparison of F1-score for the replacement of the local context we use. It can be seen that whether using global or local context, the results obtained by using the representations through the interactive fusion representation module are better than using the original representations generated by BERT. On the CoNLL04 dataset, using the interactive global context increased the F1-Score for relation extraction by 2.62\% compared to using the original BERT global context, and the results using the original BERT local context decreased the F1-Score by 1.99\%. The same pattern was observed on the CH-DDI dataset. This once again proves the effectiveness of our proposed interactive fusion representation module. We also processed the local context using self-attention mechanism and compared the results. From Table 5 and Table 6, it can be seen that the results of using attention mechanism are not significantly improved from those of using max-pooling. Considering the efficiency of the model, we ultimately decide to use max-pooling. In addition, we can observe that on the CoNLL04 dataset, the performance of relation extraction fluctuates greatly after not using the local context. For example, when using global context obtained by BERT and the token CLS respectively, the F1-score decreases by 7.77\% and 5.93\%. On the dataset CH-DDI, there is not much change in performance, which is determined by the language characteristics. English often determines relations based on certain keywords, so the introduction of complete semantic information is equivalent to introducing too much redundant information, while some complex Chinese relations need to be judged based on complete semantics. 
\begin{figure}[htbp]
\centerline{\includegraphics[width=0.7\textwidth,height=5cm]{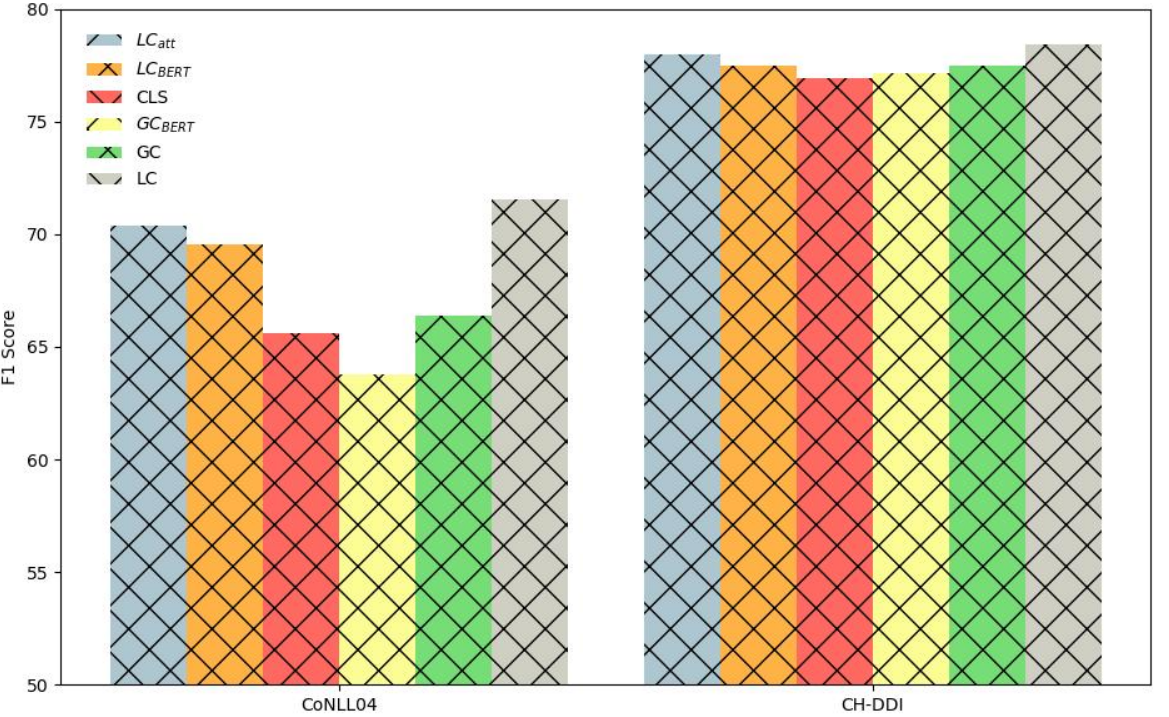}}
\caption{The comparison of F1-Score for the ablation study}
\label{fig}
\end{figure}
\section{Interpretability Analysis}
\begin{CJK*}{UTF8}{gbsn}
In this section, we visualized and analyzed the Cross Attention we use. Specifically, we aggregate the weight matrices obtained through scaled dot-product attention at the token level and normalize them. Figure 7 and Figure 8 are specific visualization examples for ConLL04 and CH-DDI. From the figures, we can see that when using the relation representations as the query, the updated representations pay more attention to the overall semantics, especially the potential entities. The example in CoNLL04 dataset shows that the model assigns higher attention to ``$Wilkes Booth$'' and ``$Lincion$'' and in the CH-DDI dataset, ``注射用环磷腺苷葡胺'' has high weight. While using entity representations as the query, Words related to relations are given higher weight. For example, in Figure 7, ``$assassinate$'' is given the highest weight and it often implies a ``$kill$'' relation in the sentence. In Figure 8, the ``禁与'' has the highest weight and it means that there may be a ``$taboo$'' relation.
\end{CJK*}
\vspace{-5mm}
\begin{figure}[htbp]
\centerline{\includegraphics[width=0.9\columnwidth,height=5cm]{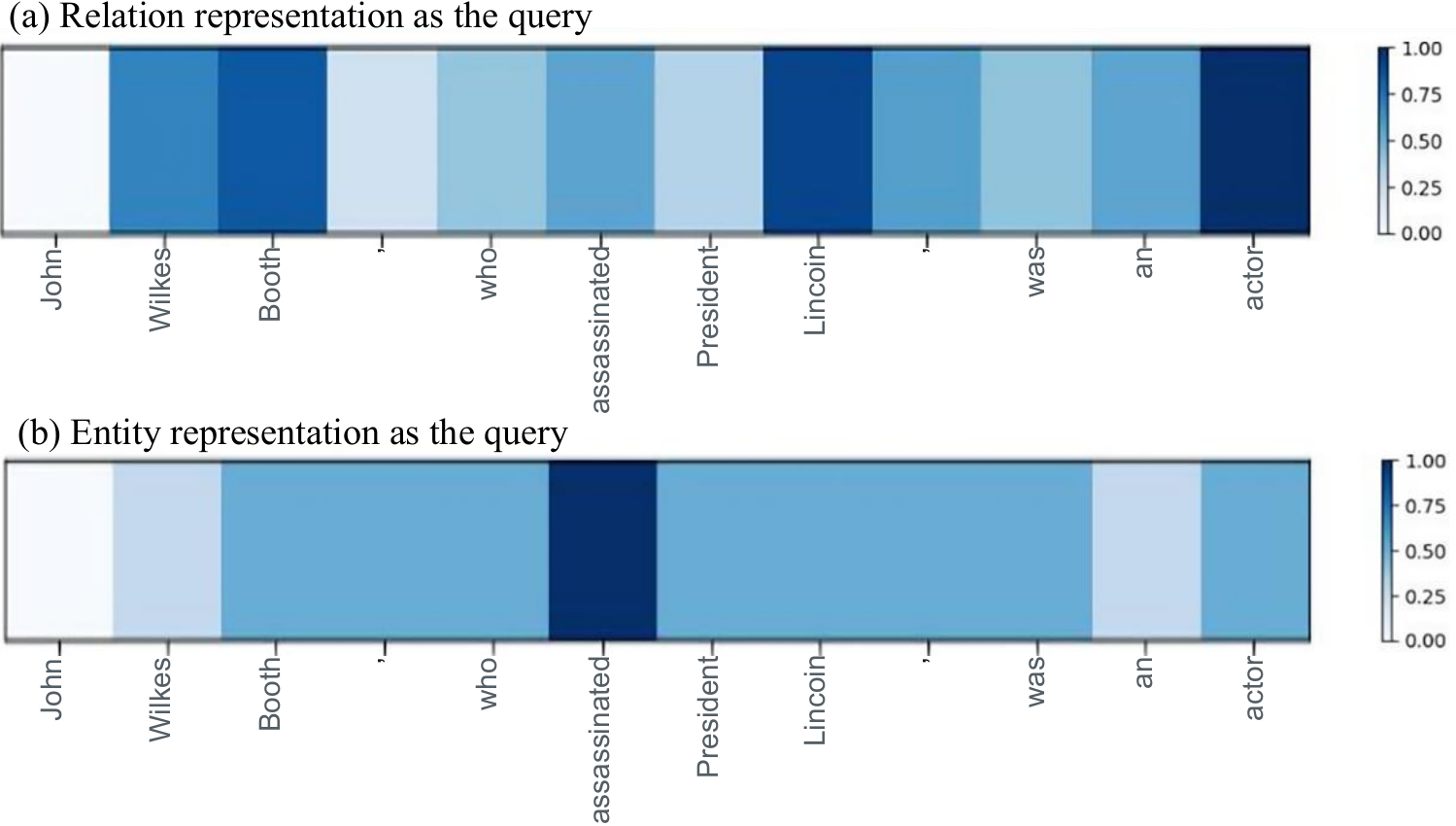}}
\caption{Visualization of Cross Attention (CoNLL04)}
\label{fig}
\end{figure}
\vspace{-5mm}
\begin{figure}[htbp]
\centerline{\includegraphics[width=0.9\columnwidth,]{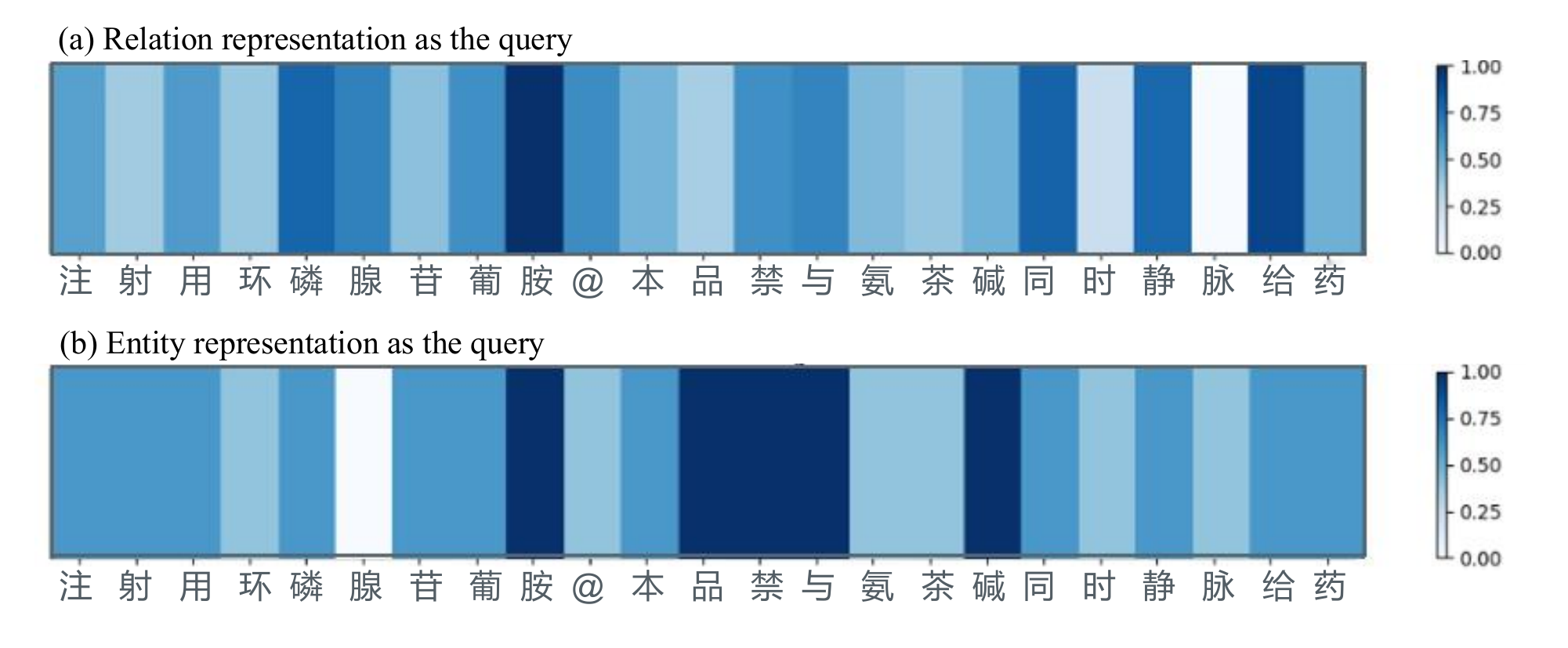}}
\caption{Visualization of Cross Attention (CH-DDI)}
\label{fig}
\end{figure}
\vspace{-5mm}
\section{Conclusion}
This paper proposes a joint entity-relation extraction model based on span and Interactive fusion representation for Chinese medical texts with complex semantics, which achieves bidirectional information interaction between entity recognition and relation extraction. When performing entity recognition, the proposed SEA module can effectively obtain contextual semantics for partitioned spans, thereby improving the performance. The local context is used to supplement semantics in relation extraction to help circumvent potential confusion within the relation classifier caused by relation overlap issues. Our model performs well on the public dataset CoNLL04 and the self-bulit CH-DDI dataset. 

In the future, we will do our best to improve our dataset CH-DDI to address the issue of data imbalance, and we will also explore more methods for information exchange in entity recognition and relation extraction.

\section{Acknowledgment}
This work was funded by the National Science and Technology Major Project (2021ZD0111000) and Henan Province Science and Technology Research Project (232102311232).
\bibliographystyle{unsrt} 
\nocite{*}
\bibliography{samplepaper}


\end{document}